# Probabilistic Arc Consistency: A Connection between Constraint Reasoning and Probabilistic Reasoning

**Michael C. Horsch**
mhorsch@cs.sfu.ca

**William S. Havens**
havens@cs.sfu.ca

Intelligent Systems Laboratory,
School of Computing Science,
Simon Fraser University,
Burnaby, B.C., Canada V5A 1S6

## Abstract

We document a connection between constraint reasoning and probabilistic reasoning. We present an algorithm, called *probabilistic arc consistency*, which is both a generalization of a well known algorithm for arc consistency used in constraint reasoning, and a specialization of the belief updating algorithm for singly-connected networks. Our algorithm is exact for singly-connected constraint problems, but can work well as an approximation for arbitrary problems. We briefly discuss some empirical results, and related methods.

## 1 INTRODUCTION

Constraint reasoning is about finding configurations which satisfy constraints, possibly optimizing the configurations according to some objective function. One of the most important tools in constraint reasoning is the process of arc-consistency, which reduces the configuration space to those configurations which meet minimal local consistencies and their immediate consequences.

In this paper we present an algorithm, called *probabilistic arc consistency* (pAC), which was developed to compute solution probabilities, *i.e.*, the frequency with which a variable takes on a particular value in all solutions, in constraint satisfaction problems. This information can be used as a heuristic to guide constructive search algorithms: for a given variable, choose the value which appears in the most solutions. Similar proposals for counting solutions or estimating solution probabilities have been made [6, 14, 4, 11, 16, 15]. Solution probabilities are orthogonal to preference over solutions (*e.g.*, [3, 1]), or probabilistic constraints (*e.g.*[1]) in which there is uncertainty regarding whether a constraint applies.

The main purpose of this paper is to document a connection between constraint reasoning and probabilistic reasoning. We show that pAC algorithm is a generalization of the basic arc consistency algorithm AC-3 [10], and is also a specialization of the belief propagation algorithm [8] for singly-connected Bayesian networks. However, since the value of our method must be established empirically, we will briefly describe some of our empirical results in Section 6. We feel our results are positive: we can report a dramatic decrease in search costs, *i.e.*, number of backtracks, using pAC as compared to related methods for counting solutions. A detailed description of our results is found in [7].

This paper is organized as follows. In Section 2, we will provide a brief description of the belief propagation algorithm for singly connected Bayesian networks, so that the comparison between algorithms can be self-contained. Section 2.1 presents an independence assumption which allows CSPs to be represented compactly, and show how this assumption changes the belief propagation algorithm. Section 3 will give a brief overview of constraint satisfaction problems, and show that the arc consistency algorithm is a specialization of belief propagation. In Section 4, we present the pAC algorithm itself, give a formal statement of correctness, and show how it generalizes the arc consistency algorithm, and specializes the belief propagation algorithm. In Section 5 we discuss the relationship between pAC and similar methods in the literature. Section 6 provides a summary of our empirical evaluation. In Section 7 we close with a discussion of these results.

## 2 BELIEF PROPAGATION IN BAYESIAN NETWORKS

Kim and Pearl [8, 13] developed a polynomial-time algorithm for singly-connected Bayesian networks. The method is based on message passing, and there are two message types: causal messages, denoted by the symbol $\pi$, are passed along the direction of the arcs in the DAG; diagnostic messages, denoted by the symbol $\lambda$, are passed against the direction of the arcs in the DAG.

The posterior probability of a variable $X$ given evidence $E$

is computed by combining the messages it receives from its parents and children:

$$P(X = x|E) = \alpha \lambda(x) \pi(x)$$

where $\alpha$ is a normalization constant.

The effect of evidence in the descendants of $X$ is summarized by:

$$\lambda(x) = \prod_j \lambda_{Y_j}(x)$$

where the quantity $\lambda_{Y_j}(x)$ summarizes the effect of evidence through child $Y_j$.

The effect of evidence in the ancestors of $X$ is summarized by:

$$\pi(x) = \sum_{U_1,\ldots,U_n} P(x|u_1,\ldots,u_n) \prod_i \pi_X(u_i) \quad (1)$$

Here, the sum is over the space of assignments to $X$'s parent variables, and the product is the product of all messages received from any parents. The quantity $\pi_X(u_i)$ summarizes the evidence through parent $U_i$.

Once $X$ has determined its own posterior probability given the evidence "mentioned" in the messages received by $X$, it passes messages to its neighbours, with care taken to avoid double-counting of evidence. The message $X$ sends to its parent $U_k$ reflects all evidence seen by $X$ except for the evidence already seen by $U_k$:

$$\lambda_X(u_k) = \alpha \sum_X \lambda(x) \sum_{U_i: k \neq i} P(x|u_1,\ldots,u_n) \prod_{k \neq i} \pi_X(u_i)$$

where $\alpha$ is a normalization constant. Notice that in this equation, $\lambda_X(u_k)$ omits any information received from $U_k$.

The message $X$ sends to its child $Y_j$ reflects all evidence seen by $X$ except for the evidence already seen by $Y_j$:

$$\pi_{Y_j}(x) = \alpha \left[ \prod_{i \neq j} \lambda_{Y_i}(x) \right] \pi(x)$$

where $\alpha$ is a normalization constant. The first product in the above expression is over all children except the one to which $X$ is sending the message; the outgoing $\pi$-message includes information received by all parents, as collected in $\pi(x)$.

The initial conditions of the algorithm are as follows: if a node $X$ has no parents, $\pi(x) = P(X = x)$. If a variable $X$ has no children, then $\lambda(x_l) = 1$ for all values $x_l \in \Omega_X$. If $X = x$ is given as evidence, then $\lambda(x) = 1$ and $\lambda(x_l) = 0$ for all values $x_l \in \Omega_X$, $x_l \neq x$.

The correctness of this algorithm is guaranteed by the fact that the sources of evidence are independent. The complexity of the algorithm, in terms of the number of messages sent, is linear in the number of nodes in the network. These results are proven in [13].

## 2.1 A SIMPLE MODEL OF CAUSAL INDEPENDENCE

*Conditional* independence is a simplifying assumption based on structure in the factorization of a joint probability distribution. This assumption is the basis for modelling joint distributions using a Bayesian network: a variable is conditionally independent of its non-descendants, given an assignment of values to the variable's parents.

*Causal* independence is a simplifying assumption based on structure in the factorization of a conditional probability distribution. The idea behind causal independence is that, for a given configuration of a subset of its parents, the probability distribution of a variable may be independent of the remainder of its parents. To take an example from a diagnostic model, the proposition that a car starts when the ignition key is turned may depend directly on factors such as the amount of gas in the tank, the state of the battery, *etc.*. The conditional probability table allows for arbitrarily complex interactions among the parent values. However, the domain expert may judge that a car cannot start if the battery is dead, no matter how much gas is in the tank. On the other hand, if the battery voltage is low or high, the probability of starting may depend on the quantity of gas in the tank.

There have been many formalizations of causal independence (*e.g.*, [13, 17]), which allow for many interesting and important variations. The assumption we will use here is very simple. Suppose a variable $X$ has parents $\{U_1, \ldots, U_n\}$. We will assume that the conditional probability table $P(X|U_1, \ldots, U_n)$ can be factored as follows:

$$P(X|U_1, \ldots, U_n) = \alpha \prod_{i=1}^n P(X|U_i)$$

where $\alpha$ is a normalizing constant. We will make this assumption for every variable in a singly-connected Bayesian network.

Under this assumption, the $\pi$-terms in the polytree algorithm are composed as follows:

$$\pi(x) = \alpha \prod_i \sum_{U_i} P(x|u_i) \pi_X(u_i) \quad (2)$$

The $\pi$-message sent from $X$ to child $Y_j$ is unchanged, once $\pi(x)$ is computed.

The $\lambda$-message sent to parent $U_k$ is as follows:

$$\lambda_X(u_k) = \alpha \sum_X \lambda(x) \prod_{k \neq i} \sum_{U_i} P(x|u_i) \pi_X(u_i)$$

Note that the $\lambda$-message is sent to $U_i$ by summing over $X$'s values. This is done because the algorithm assumes that the conditional distribution $P(X|U_1, \ldots, U_n)$ is stored locally for $X$, but is not stored locally for $U_k$.

Given our assumption of causal independence for every variable in the network, the low dimensionality of $P(X|U_i)$, and the fact that $P(X|U_i)P(U_i) = P(U_i|X)P(X)$, it is not unreasonable to allow $U$ to store local information $P(U_i|X)$, even when $X$ stores $P(X|U_i)$. We assume that this information is available. In this case, $X$ need not sum out its own values before sending the $\lambda$-message; it can let $U_i$ sum out over $X$'s values, using $P(U_i|X)$. Thus, we can simplify the $\lambda$-messages by sending the following $\lambda'$-message from $X$ to $U_i$:

$$\lambda'_{U_i}(x) = \alpha \lambda(x) \prod_{k \neq i} \sum_{U_i} P(x|u_i) \pi_X(u_i)$$

This message must be "interpreted" by $U_i$ by summing out the values of $X$. Under our causal independence assumption, the original $\lambda$-message can be reconstructed by $U_i$ using the following:

$$\lambda_X(u_i) = \alpha \sum_{x \in X} P(u_i|x) \lambda'_{U_i}(x)$$

Note that this is exactly how $\pi$-messages are treated, according to the causal independence assumption. This implies that under our assumption, the parent-child relationship is symmetric, and we need only pass one kind of message (to distinguish this method from the more general case where our causal independence assumption does not hold, we will call them $\gamma$-messages).

The message to neighbour $V_i$ is simply the product of all the information received by $X$ except for the information received from $V_i$ itself. We define

$$Bel(x) = \alpha \prod_i \lambda_{V_i}(x)$$

As above, $\lambda_{V_i}(x)$ describes the information about $X$ received from $V_i$, but is not available directly. It is computed from the $\gamma$-message sent from $V_i$ to $X$ as follows:

$$\lambda_{V_i}(x) = \alpha \sum_{v \in V_i} P(x|v) \gamma_X(v)$$

The outgoing message to all neighbours is:

$$\gamma_{U_i}(x) = \alpha \prod_{k \neq i} \sum_{V_i} P(x|v_i) \lambda_{V_i}(x)$$

We have seen how a simple assumption of causal independence leads to a simplification of the polytree belief propagation algorithm. We have replaced $\pi$ and $\lambda$ messages with $\gamma$ messages. The determination of the local quantity $\lambda_V$ used above is derived from $\gamma_V$, and it represents the information received by $X$ from neighbour $V$.

## 3 CONSTRAINT SATISFACTION PROBLEMS

A constraint satisfaction problem (CSP) is posed as a set of variables, a domain for each variable, and a set of constraints over tuples of domain values (readers who would like more background on CSPs than can be presented in here are referred to [9]). The problem is to find an assignment of values to all variables which satisfies all the constraints. We will denote variables $X_i, 0 \leq i < n$, and the domains $D_i$. We will focus on binary CSPs, in which all the constraints are sets of tuples from pairs of domains, $C_{ij} \subset D_i \times D_j$. As well, we will limit our discussion to domains which are finite and discrete.

One approach to solving CSPs is to use constructive search. Values are assigned to variables in some sequence, and after each assignment, the assignment is checked for consistency with the constraints. If the assignment is consistent, the search continues recursively. If an assignment is inconsistent, a different value is chosen, and if no values for the current variable in the sequence are consistent, the search backtracks to the previously assigned variable, and continues the search. Heuristic information can be used to order the values and the variables in an attempt to speed up search.

There is a well-documented problem with simple constructive search, namely that there are values, and combinations of values, which never appear in any solution. If the constructive search method tries one of these assignments, it is guaranteed to have to backtrack. Furthermore, unless special care is taken in the search algorithm, these assignments may be tried often during the search. These problems have been well-studied [10], and several pre-processing algorithms have been devised to limit their effects. In particular, an algorithm called AC-3 reduces the domains of the variables in a CSP, by removing those values which are not consistent with some value of its neighbouring variables. We present AC-3 in more detail in the following section.

### 3.1 ARC CONSISTENCY

Arc consistency can be defined as follows [10]. An arc (or edge) $C_{XY}$ in a constraint graph for a binary CSP is arc consistent iff for all $x \in D_X$, there is a value $y \in D_Y$ such that $(x, y) \in C_{XY}$. In this definition, we assume that there are no unary constraints which may rule out values in $D_X$ or $D_Y$. Clearly, CSPs need not be arc consistent, but if the domains of the CSP variables were reduced such that each arc was arc consistent, the cost of search could be reduced. When all arcs are arc consistent, we say the problem is arc consistent.

An algorithm for computing arc consistency is given in Figure 1 and is due to [10]. The underlying idea is to cycle through the variables in some order, and reduce the domain of each variable such that only values which are arc consistent remain. The effects of reducing a domain are *propagated* to neighbouring variables.

There are two key aspects of the AC-3 algorithm. First,

in Revise, a domain element is removed if there is some neighbour that does not support the element. Equivalently, a domain element is supported if there is a supporting element in all neighbours' domains. Let $F_i \subset D_i$ be the set of supported values for variable $i$.

$$F_i = \{x \in D_i | \forall j \in N_i, \exists y \in D_j : (x,y) \in C_{ij} \land y \in F_j\}$$

where $N_i$ are the neighbours of $X_i$, i.e., variables which share a constraint with $i$, and $F_j \subset D_j$ is the set of supported values for variable $j$. We will write $F_i(x)$ to represent the proposition $x \in F_i$, and similarly $C_{ij}(x,y)$ to represent $(x,y) \in C_{ij}$. Thus we have the following equivalence

$$F_i(x) = \forall_{j \in N_i} \exists_{y \in D_j} (C_{ij}(x,y) \land F_j(y))$$

The use of $\forall$ and $\exists$ is purely by convention; using $\times$ and $+$ to represent $\land$ and $\lor$, resp. and using $\sum$ and $\prod$ to represent boolean sums and products, we can rewrite this equivalence as follows:

$$F_i(x) = \prod_{j \in N_i} \sum_{y \in D_j} (C_{ij}(x,y) \times F_j(y)) \qquad (3)$$

This is the domain update rule applied to each variable during the propagation of arc consistency.

The second key element in AC-3 is the mechanism for cycling through the variables to ensure soundness of the algorithm. As shown, a queue of pairs of constrained variables is maintained, and pairs are removed and added to the queue.

The purpose of the queue is to guarantee that changes to the domain of one variable are propagated to its neighbours. In other words, selecting edge $(k,m)$ from the queue results in a call to Revise$(k,m)$, which updates $F_k$, as in Equation 3 above, using $F_m$. The queue contains and orders messages between variables, and messages are processed in a sequential manner. The propagation mechanism could equally well be expressed as a distributed algorithm: after each change in the domain of a variable, the variable's neighbours are made aware of the changes, and change their domains accordingly.

Note also that when the processing of edge $(k,m)$ results in a domain reduction (i.e., Revise$(k,m)$ returns true), arcs $(i,k)$ are added to the queue, but the arc $(m,k)$ is explicitly omitted from the queue. That is, when a variable's domain is reduced due to information received from a neighbour, it does not cause the neighbour to be checked for revision. This is a matter of efficiency only, because, due to the symmetric nature of a binary constraint, the neighbour's domain cannot be reduced if the arc were included in the queue.

Thus we have shown informally that AC-3 can be expressed as a special case of belief propagation in singly-connected

```
procedure Revise(i, j):
  begin
    delete := false
    for each x ∈ D_i do
      if there is no y ∈ D_j such that (x,y) ∈ C_ij then
        remove x from D_i
        delete := true
      end if
    end for
    return delete
  end

procedure AC-3
  begin
    Q := {(i,j)|(i,j) ∈ arcs(G), i ≠ j}
    while Q not empty do
      select and delete any arc (k, m) from Q;
      if Revise(k,m) then
        Q := Q ∪ {(i,k)|(i,k) ∈ arcs(G), i ≠ k, i ≠ m}
      end if
    end while
  end
```

Figure 1: AC-3. An algorithm for achieving arc consistency.

networks. To make the case explicit, we first observe that in a binary CSP, the neighbours of a node are causally independent, in the sense described in the previous section. Second, we replace the conditional probability distributions $P(X_i|X_j), i \neq j$ with boolean arrays representing $C_{ij}$. Finally, the operations $+$ and $\times$ are replaced by boolean operations $\lor$ and $\land$, respectively.

## 4 PROBABILISTIC ARC CONSISTENCY

In this section we present the pAC algorithm, which we use to compute (or approximate, as we will see) the solution probabilities for every variable in a CSP. This information can be used as a heuristic to guide constructive search algorithms: e.g., for a given variable, choose the value which appears in the most solutions.

**Definition 1** *Let $<V, B>$ be a binary constraint satisfaction problem with variables $V$, and binary constraints $B$, such that the constraint graph $G = <V, B>$ is singly-connected. For every $X \in V$, let $D_X$ be the domain of $X$, i.e., $D_X = \{x_1, \ldots, x_m\}$. For every pair of variables $(X, Y)$ such that there is a binary constraint $C_{XY} \in B$, define the following:*

$$C_{XY}(i,j) = \begin{cases} 1 & \text{if } (x_i, y_j) \in C_{XY} \\ 0 & \text{otherwise} \end{cases}$$

$$M_{XY}^{(0)}(i) = 1$$

$$S_{XY}^{(k)}(i) = \sum_{j=1}^{m} C_{XY}(i,j) M_{XY}^{(k)}(j)$$

$$F_X^{(k)}(i) = \alpha \prod_{\{Y|(X,Y)\in B\}} S_{YX}^{(k)}(i)$$

$$\text{where} \quad \alpha \quad \text{is such that} \sum_{i=1}^{m} F_X^{(k)}(i) = 1$$

$$M_{XY}^{(k+1)}(i) = \begin{cases} \dfrac{F_X^{(k)}(i)}{S_{YX}^{(k)}(i)} & \text{if } S_{YX}^{(k)}(i) > 0 \\ 0 & \text{otherwise} \end{cases}$$

The interpretation of the quantities $S_{XY}^{(k)}(i)$, $F_X^{(k)}(i)$, $M_{XY}^{(k+1)}(i)$ is as follows. The binary constraints in $B$ are expressed numerically in $C_{XY}$, where $C_{XY}(i,j) = 1$ iff the pair $(x_i, y_j)$ satisfies the binary constraint. The interpretation of the remaining quantities is the subject of the following theorems (the proofs appear in [7]).

**Lemma 1** *Let $X, Y$ be any constrained pair of variables in a CSP $G$ whose constraint graph is singly-connected. Let $G'$ be the subproblem constructed in the following way: Include in $V'$ only those variables from $V$ which are $k$ steps away from $Y$ in the constraint graph of $G$, and which are separated from $Y$ by $X$. Likewise, include in $B'$ only those constraints $C_{XY} \in B$ for which both $X \in V'$ and $Y \in V'$. $M_{XY}^{(k)}(i)$ is proportional to the number of times value $x_i \in D_X$ is used in all solutions of the sub-problem $G'$.*

**Lemma 2** *Let $X$ be any variable in a CSP $G$ whose constraint graph is singly connected. $F_X^{(k)}(i)$ is the relative frequency of the use of $x_i \in D_X$ in all solutions of the sub-problem including only those variables in $G$ which are distance $k$ or less from $X$. If $G$ is unsatisfiable, $F_X^{(k)}(i) = 0$ for all $i$.*

**Theorem 3** *Let $G = <V, B>$ be a CSP such that the constraint graph for $G$ is singly-connected, with diameter $d$. For any variable $X \in V$, the relative frequency that value $x_i \in D_X$ is used in all solutions of $G$ is $F_X^{(d)}(i)$.*

The pAC equations can be expressed as a distributed procedure. Each variable $X_i$ is initialized to have uniform distributions. At each time step, each variable saves its previous distribution ($F$), and prepares to handle incoming messages. Messages ($M$) from neighbouring variables are processed ($S$), and the results are stored locally, so that messages need not be sent to all neighbours if no changes were made in the distribution. The new distribution is computed by forming the product of all information stored from the most recent message received from all neighbours. Finally, if the variable's distribution has changed significantly, a new message ($M$) is sent to all neighbours, taking care not to double count.

The pAC equations require arbitrary precision floating point numbers. In our implementation, we use 64 bit floating point numbers, at the risk of a non-trivial loss of precision when the CSPs get very large. In our experimental work, we can observe this phenomenon in about 5% of the large problems we have tried.

The pAC equations generalize arc consistency. If boolean values are used instead of probabilities, and boolean operators "and" and "or" instead of floating point multiplication and addition, the algorithm computes arc consistency. In singly-connected CSPs, the arc consistent domain values are used in some solution, and therefore have non-zero solution probability.

The pAC equations are also a special case of Pearl & Kim's belief propagation algorithm for singly-connected belief networks. To prove this informally, it is sufficient to observe the following correspondences: $C_{XY}(i,j)$ can be represented as $P(X = x_i|Y = y_j)$, under the causal independence assumption given in Section 2.1; $S_{XY}^{(k)}(i)$, $M_{XY}^{(k)}(i)$, and $F_X^{(k)}(i)$ correspond to $\lambda_Y(x_i)$, $\gamma_Y(x_i)$, and $Bel(x_i)$, resp., after $k$ messages were received by $X$.

The guarantee of correctness only holds in CSPs with singly connected constraint graphs. For more general CSPs, the equations can be used to approximate solution probabilities for these problems, by iterating the equations some number of times. This approach has parallels with relaxation methods for belief propagation in Bayesian networks [12], and decoding turbo-codes [5]. The method is not guaranteed to converge to a stable set of probability distributions, and if it does, there is no guarantee that the approximations are useful. Thus the value of the method is an empirical question, which we explore elsewhere [7], and summarize in Section 6.

To limit computation costs, we use two parameters, $\epsilon$ and MaxIter, to detect convergence or non-convergence. Iteration continues while both of the following conditions are true:

$$\max_X \sum_i (F_X^{(k+1)}(i) - F_X^{(k)}(i))^2 > \epsilon \qquad (4)$$

$$k < \text{MaxIter} \qquad (5)$$

When the change in solution probability is less than a given $\epsilon$, the process is declared to have converged; when the number of iterations exceeds MaxIter, the process is halted without convergence.

The approximate solution probabilities computed using pAC iteratively can be used to guide constructive search. For example, approximate solution probabilities can be computed before each assignment, and the probability used to induce a variable ordering, *i.e.*, choose the variable whose maximum probability domain value is maximum over all unassigned variables, as well as a value ordering *i.e.*, choose the most likely domain value for the variable.

## 5 RELATED WORK

Probabilistic arc consistency is strongly related to techniques proposed in the literature. Dechter and Pearl's [4] single spanning tree method (SST) determines an exact solution count in singly-connected CSPs. The variables in a CSP are given an arbitrary ordering, in which $X_j$ is chosen as the root. The number of solutions in which $X_j$ takes value $v_t$ is computed recursively by the following:

$$N(X_j = x) = \prod_{C'} \sum_{D'} N(X_c = y)$$

where

$$C' = \{c | X_c \text{ is a child of } X_j\}$$
$$D' = \{y \in D_c | (x, y) \in C_{jc}\}$$

The leaf variables $X_l$ in the ordering have $N(X_l = v) = 1$ for all $v \in D_l$. This method must be repeated for each variable if solution counts are necessary for all variables. For more general CSPs, the method is applied to the subproblem consisting of the tightest constraints in the original problem. The method is efficient, but can result in over-optimistic approximations for the number of solutions in the original problem.

Meisels et al.'s [11] universal propagation method (UP) uses a Bayesian network to compute solution counts for a designated variable, making the same independence assumptions in Section 2.1. The Bayesian network is the constraint graph whose constraints are directed according to a pre-established variable ordering. The goal of the computation is to find the marginal probability distribution $P(X_j)$ for the designated sink variable $X_j$. Under the same assumption of causal independence as described in Section 2.1:

$$P(X_j = x) = \alpha \prod_{C'} \sum_{D'} P(X_j = x | X_c = y) P(X_c = y)$$

where

$$C' = \{c | X_c \text{ is a parent of } X_j\}$$
$$D' = \{y \in D_c\}$$

and where $P(X_j | X_c)$ represents the constraint between $X_j$ and $X_c$, and $\alpha$ is a normalization constant. When applied to CSPs whose constraint graph is singly-connected, the method reduces to that of [4], with the addition of a normalization constant, and computes exact solution probabilities for a designated variable. For more general CSPs, the same method is applied, producing approximate solution counts, which can be over-optimistic.

Vernooy and Havens [16] method extends [4] by constructing a forest of spanning trees from the original CPS's constraint graph. Using either of the previous methods, the solution counts for each tree are determined. An approximate solution count for the original CSP is determined by combining the results from each tree, under the assumption that the solutions to each tree are independent. Using this assumption, an approximation is made for $P(X_i)$, that is, the distribution of solutions for variable $X_i$ in the original CSP, by normalizing the product of the distributions from all trees:

$$P(X_i = x) = \alpha \prod_k P^{(k)}(X_i = x)$$

where $\alpha$ is a normalization constant, and $P^{(k)}(X_i = x)$ is the solution probability derived from the $k$th spanning tree. Vernooy and Havens explore this multiple spanning tree method (MST) as both a static and dynamic heuristic.

Peleg [14] develops a probabilistic relaxation method intended to find a satisfying assignment to the variables in a CSP. Although different in intent, the algorithm itself is similar to the algorithms discussed above. The difference, apart from the motivation, is a single factor in the expression for $F_X^{(k)}(i)$ in the pAC algorithm. Peleg's formula uses the previous iteration's estimate to modify the current estimate:

$$F_X^{(k)}(i) = \alpha F_X^{(k-1)}(i) \prod_{\{Y | (X,Y) \in B\}} S_{YX}^{(k)}(i)$$

It should be noted that in this interpretation, $F_X^{(k)}(i)$ is no longer a solution probability, in the sense used in this paper. Rather, Peleg's method is intended to converge to a solution to the CSP, i.e., a distribution in which every variable has only one value with non-zero "probability."

The work of Shazeer et al. [15] on computing probabilistic preferences over solutions to CSPs develops an algorithm which is essentially the same as the pAC algorithm. The main difference is that pAC uses the constraint graph structure itself, whereas the proposal by Shazeer et al. unrolls the constraint graph to form a large singly-connected structure in which the nodes and arcs are repeated some fixed number of times. A minor difference is that pAC assumes a uniform prior over all values in a domain, so that the resulting distributions are solution probabilities; the proposal by Shazeer et al. allows arbitrary probability distributions over the domain values to express more or less likely values. Thus these distributions are not strictly solution probabilities, in the sense used in this paper.

## 6 RESULTS

We have conducted an extensive investigation into the performance of pAC as a method for approximating the solution frequencies of binary CSPs [7]. We evaluated the accuracy of the approximation computed by pAC, and examined its effectiveness as a heuristic in constructive search.

We were able to show that there was a high correlation between the exact solution probabilities (as computed by exhaustive search) and the approximations computed by pAC. We were also able to show that when used as a variable ordering heuristic, the approximate solution probabilities computed by pAC substantially reduce backtracking in constructive search, by up to two orders of magnitude. Space constraints prohibit a complete presentation of these results here.

Our experiments were performed on randomly constructed CSPs, which can be described using four parameters: the number of variables, $n$, the number of values $m$, the constraint density $p_1$ (the probability that any two variables share a constraint), and the constraint tightness, $p_2$ (the probability that any pair of values is disallowed in a given constraint). It is well known that random CSPs vary greatly in difficulty, across the parameter space for these problems.

We evaluated the accuracy of the approximation by direct comparison to exact solution counts, computed by exhaustive search. We applied pAC to 3 sets of random CSPs of varying topology and constrainedness, with $n = 20, m = 10$. The difficulty of these problems varied from very simple (having very many solutions, or being obviously over-constrained), to very difficult (having only only a few solutions). The convergence criteria were set fairly high, with $\epsilon = 10^{-5}$ and MaxIter= 1000.

Of the problems with fewer than 1 million solutions and at least one solution, the average correlation between the exact solution probability and the approximation determined by pAC ranged between 0.83 for for $p_1 = 0.2$ (sparse graphs), through 0.78 for $p_1 = 1.0$ (complete graphs). As well, pAC was able to identify the majority of over-constrained problems very quickly; this is what should be expected of a method which generalizes arc consistency.

For about 10% of the problems, pAC failed to converge. In some cases, the convergence was just very slow, and in other cases, the values computed by pAC oscillated between two distinct points in the distribution space. When pAC does not converge, any degree of accuracy is possible.

We also evaluated the effectiveness of the heuristic as used in constructive search. We compared pAC to the methods of Dechter and Pearl (SST) [4], Meisels et al. (UP) [11], Vernooy and Havens (SMST) [16] and Peleg [14]. Each of these methods was used as a static value ordering heuristic (i.e., the heuristics were computed as a preprocessing step, but not updated as assignments were made). The MST method of Vernooy and Havens was also used dynamically (DMST), i.e., the solution probabilities were recomputed after every assignment. A random value ordering strategy was included in this comparison to provide a baseline.

We applied simple constructive search to 218 random problems, using the various heuristics as a value ordering heuristic. These problems were selected from a larger set of random problems, from which we discarded the over-constrained problems. They were constructed with 20 variables and 10 values each, and $p_1 = 1.0$ (complete graphs), and $p_2 \in [0.1, 0.23]$. Some of these problems were very easy, but many were much more difficult.

Figure 2 shows the results. The horizontal axis shows search costs, in terms of the number of backtracks; the vertical axis indicates the cumulative fraction of the problems solved. Each curve shows the cumulative fraction of problems solved using a given number of backtracks. For example, the median number of backtracks for each method can be read from the graph on the horizontal line through the 0.5 mark.

The graph clearly shows that pAC is superior to all the methods, except for Peleg's method, which was able to converge on a solution for roughly 66% of the problems, requiring no backtracking whatsoever. However, Peleg's method is not much of an improvement over random value orders for some problems. We repeated this experiment for random problems with different constraint densities, with very similar results.

We also evaluated the approximate solution probabilities computed by pAC when used as a dynamic variable and value ordering heuristic, i.e., after each assignment, the solution probabilities were recomputed. We found that the reduction in search costs is dramatic, up to two orders of magnitude smaller than First Fail [6] or Least-Constrained [2].

The price for this success is the cost of computing the heuristic values: even for relatively large $\epsilon$, say $\epsilon = 0.1$, and few iterations, e.g., MaxIter= 50, the cost of performing all the floating point operations is such that search requires almost an order of magnitude more time using approximate solution probabilities than the First Fail or Least-Constrained heuristics.

## 7 DISCUSSION AND FUTURE WORK

In this paper we have presented three algorithms, belief propagation (BP) in singly-connected Bayesian networks, arc consistency (AC-3) for binary constraint satisfaction problems, and probabilistic arc consistency (pAC) for binary CSPs. We have shown that probabilistic arc consistency is a special case of the belief propagation algorithm, under an assumption of causal independence. We have also shown that the arc consistency algorithm is a special case of probabilistic arc consistency, specialized to perform of boolean arithmetic.

In the case of singly-connected topologies, all three algorithms are exact, and run in polynomial time. The BP algorithm assumes conditional independence of a variable's parents for correctness. Likewise, when a CSP's constraint graph is singly-connected, the pAC algorithm is also exact, correctly computing solution probabilities for all variables. Finally, AC-3 is exact for all topologies and runs in polynomial time. However, when the constraint graph is singly-connected, constructive search over the reduced domains is backtrack-free. This property is not true for AC-3 on binary CSPs of arbitrary topology.

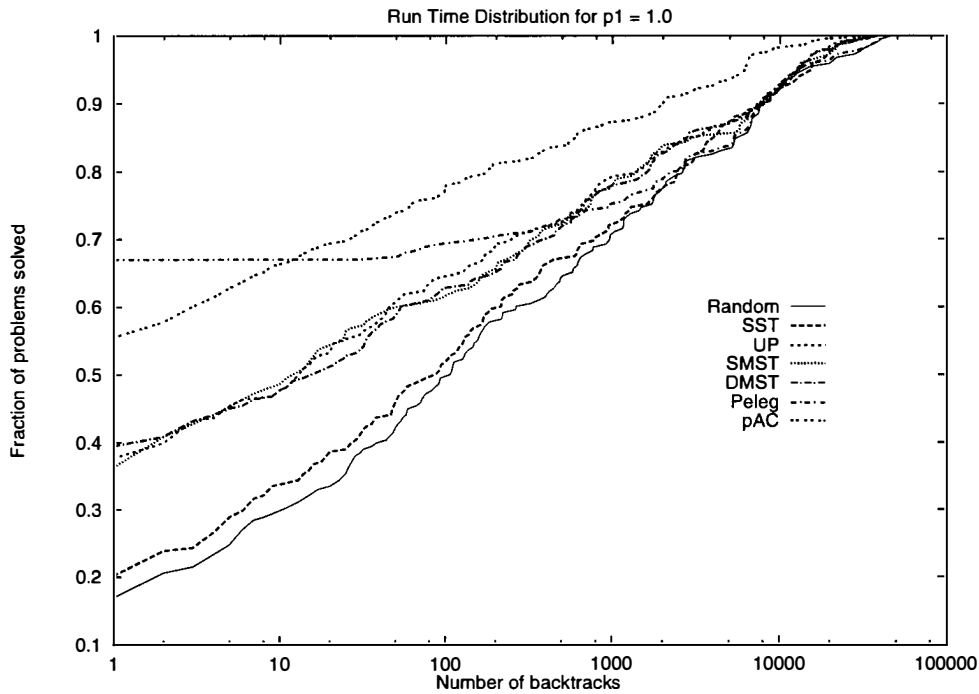

Figure 2: A plot of the number of backtracks required to solve a set of random CSPs with constraints between every pair of variables, and between 10 and 23 percent of the pairs disallowed by the constraints. The plot shows the cumulative fraction of the number of problems solved using a given number of backtracks.

The BP algorithm has been suggested and studied as an approximation method for more general topologies (e.g., [13, 12]). It is almost startling that the propagation method, designed for singly connected graphs, often converges on arbitrary graphs to a distribution which is a reasonable approximation of the exact distribution. Obviously, the method will work well in graphs in which the conditional dependence is weak.

Since approximation in Bayesian networks of arbitrary topology is NP-hard, it is expected that it might take a long time for BP to converge in at least a few instances. This has been observed in empirical studies, and in some cases, the algorithm has been observed to oscillate indefinitely [12].

The pAC algorithm can also be used to approximate solution probabilities for constraint problems of arbitrary topology. Although pAC has stopping conditions, it is not guaranteed that the process will converge. Again, as solving binary CSPs is an NP-hard problem, it was expected that convergence of PAC might be slow in some cases. In some cases, as with the BP algorithm, the iterative process oscillates.

If the process converges, it usually converges to a good approximation to the solution probabilities. However, when pAC is oscillating, any arbitrary cutoff point is likely to provide no better than random choice heuristic information. Fortunately, the oscillation problem can disappear when values are assigned; this means that oscillation will incur some extra search costs, but these seem to be small. Unfortunately, the reverse is true as well: making assignments can also induce oscillation on the remaining variables.

We have found that if pAC oscillates, it does so between two "poles," that is, two distributions which favour dramatically different configurations. We were able to construct a problem with 5 variables and two "loops," on which pAC demonstrates oscillatory behaviour. The problem in this example is that messages circulate around the loop, and due to loops having different lengths, arrive at loop junctures "out of phase."

The relationship between belief propagation, arc consistency and probabilistic arc consistency has demonstrated to be useful for constraint reasoning. In hindsight, it should have been obvious that the relationship would lead to fairly accurate approximations to solution probabilities, based on the experience of applying belief propagation to arbitrary Bayesian networks. The relationship should also prove use-ful for probabilistic reasoning in ways that may not yet be obvious.

We are currently investigating a number of interesting issues. Computing approximate solution probabilities using pAC as described in this paper is an expensive operation. There are many variations of the scheme which could be explored. The goal would be to reduce computational costs, and maintain the effectiveness of the heuristic. Obviously, the convergence criteria can be modified to reduce the number of iterations. Propagation could be limited to some subset of the constraints (the work of [4, 16] limit propagation to tree structures, but other subsets are possible). We are also investigating techniques to reason explicitly about the trade-offs between propagating solution probabilities, and search costs.

Recently, a theory of so-called "Semiring CSPs" [1] has unified several variations of constraint problems, including satisfaction of classical CSPs, and optimization in probabilistic, fuzzy and valued CSPs. As described in [14, 15], the propagation of solution probabilities can perform some kinds of optimization by giving domain elements non-uniform *a priori* weights. We are looking at using propagation techniques to provide heuristic information for optimization of different kinds of objective functions, extending both [3, 1].

### Acknowledgements

Thanks to Matt Vernooy for providing the data from his thesis, and to the anonymous referees for their helpful comments.

## References

[1] S. Bistarelli, U. Montanari, F. Rossi, T. Schieux, G. Verfaille, and H. Fargier. Semiring-Based CSPs and Valued CSPs: Frameworks, Properties and Comparison. *Constraints*, 4(3):199–240, 1999.

[2] D. Brelaz. New methods to color the verticies of a graph. *JACM*, 22(4):251–256, 1979.

[3] Rina Dechter, Avi Dechter, and Judea Pearl. Optimization in constraint networks. In *Influence Diagrams, Belief Nets and Decision Analysis*, pages 411–425. John Wiley and Sons Ltd, 1990.

[4] Rina Dechter and Judea Pearl. Network-based heuristics for constraint-satisfaction problems. *Artificial Intelligence*, 34:1–34, 1988.

[5] Brendan J. Frey and David J. C. MacKay. A revolution: Belief propagation in graphs with cycles. In *Advances in Neural Information Processing Systems*, pages 479–485, 1998.

[6] Richard M. Haralick and Gordon L. Elliot. Increasing tree search efficiency for constraint satisfaction problems. *Artificial Intelligence*, 14(3):263–313, 1980.

[7] Michael C. Horsch and William S. Havens. How to count Solutions to CSPs. Technical report, School of Computing Science, Simon Fraser University, 2000.

[8] Jin H. Kim and Judea Pearl. A computational model for causal and diagnostic reasoning in inference systems. In *Proceedings of the Eighth International Joint Conference on Artificial Intelligence*, pages 190–193, 1983.

[9] Vipin Kumar. Algorithms for constraint-satisfaction problems: A survey. *AI Magazine*, 13(1):32–44, 1992.

[10] Alan K. Mackworth. Consistency in networks of relations. *Artificial Intelligence*, 8(1):99–118, 1977.

[11] Amnon Meisels, Solomon Ehal Shimonoy, and Gadi Solotorevsky. Bayes networks for estimating the number of solutions to a csp. In *Proceedings of the Fourteenth National Conference on Artificial Intelligence*, 1997.

[12] Kevin P. Murphy, Yair Weiss, and Michael I. Jordan. Loopy belief propagation for approximate inference: An empirical study. In *Proceedings of the Fifteenth Conference on Uncertainty in Artificial Intelligence*, pages 467–475, 1999.

[13] Judea Pearl. *Probabilistic Reasoning in Intelligent Systems: Networks of Plausible Reasoning*. Morgan Kaufmann Publishers, Los Altos, 1988.

[14] Shmuel Peleg. A new probabilistic relaxation method. *IEEE Transactions on Pattern Matching and Machine Intelligence*, 2(4):362–369, 1980.

[15] Noam M. Shazeer, Michael L. Littman, and Greg A. Keim. Constraint satisfaction with probabilistic variable values. Technical Report CS-99-03, Duke University, Department of Computer Science, 1999.

[16] Matt Vernooy and William S. Havens. An examination of probabilistic value-ordering heuristics. In *Proceedings of the 12th Australian Joint Conference on Artificial Intelligence*, 1999.

[17] Nevin Lianwen Zhang and David Poole. Exploiting Causal Independence in Bayesian Network Inference. *Journal of Artificial Intelligence Research*, 5:301–328, 1996.